%% file: paper.tex
\documentclass{article}

 \usepackage[nonatbib,final]{nips_2017}

\usepackage[utf8]{inputenc} 
\usepackage[T1]{fontenc}    
\usepackage{hyperref}       
\usepackage{url}            
\usepackage{booktabs}       
\usepackage{amsfonts}       
\usepackage{nicefrac}       
\usepackage{microtype}      

\usepackage[utf8]{inputenc} 
\usepackage[T1]{fontenc}    
\usepackage{hyperref}       
\usepackage{url}            
\usepackage{booktabs}       
\usepackage{amsfonts}       
\usepackage{nicefrac}       
\usepackage{microtype}      
\usepackage{amsmath}
\usepackage{graphicx} 
\usepackage{subcaption}

\usepackage{graphicx}
\usepackage{color}

\usepackage{multirow}

\newcommand{\fig}[1]{Fig.~\ref{fig:#1}}

\newcommand{\secc}[1]{Section~\ref{sec:#1}}

\newcommand{\rulesep}{\unskip\ \vrule\ }

\newcommand{\etal}{\textit{et al}.\:}

\newcommand{\drnet}{{\sc DrNet} }

\title{Unsupervised Learning of Disentangled Representations from Video}

\author{
  Emily Denton \\
  Department of Computer Science\\
 New York University\\
  \texttt{denton@cs.nyu.edu} \\
  \And
     Vighnesh Birodkar\\
  Department of Computer Science\\
 New York University\\
  \texttt{vighneshbirodkar@nyu.edu} \\
}

\begin{document}

\maketitle
\vspace{-3mm}
\begin{abstract}
  We present a new model \drnet that learns disentangled image
  representations from video. Our approach leverages the temporal
  coherence of video and a novel adversarial loss to learn a
  representation that factorizes each frame into a stationary part and
  a temporally varying component. The disentangled representation can
  be used for a range of tasks. For example, applying a standard LSTM
  to the time-vary components enables prediction of future frames. We
  evaluate our approach on a range of synthetic and real videos,  demonstrating the ability to coherently generate hundreds of steps into the future. 
\end{abstract}
\vspace{-1mm}
\input{intro}

\input{related}

\input{approach}
\input{experiments}

\section{Discussion}
In this paper we introduced a model based on a pair of encoders that
factor video into content and pose. This seperation is achieved during
training through novel adversarial loss term. The resulting
representation is versatile, in particular allowing for stable
and coherent long-range prediction through nothing more than a
standard LSTM. Our generations compare favorably with leading
approaches, despite being a simple model, e.g. lacking the GAN losses or
probabilistic formulations of other video generation
approaches. Source code is available at \url{https://github.com/edenton/drnet}.

\small
\bibliography{bibliography} 
\bibliographystyle{ieee}

\input{supp}

\end{document}

%% file: intro.tex
\vspace{-3mm}
\section{Introduction}
\vspace{-2mm}
Unsupervised learning from video is a long-standing problem in computer vision and machine learning. The goal is to learn, without explicit labels, a representation that generalizes effectively to a previously unseen range of tasks, such as semantic classification of the objects present, predicting future frames of the video or classifying the dynamic activity taking place. There are several prevailing paradigms: the first, known as self-supervision, uses domain knowledge to implicitly provide labels (e.g. predicting the relative position of patches on an object \cite{doersch2015unsupervised} or using feature tracks \cite{wang2015unsupervised}). This allows the problem to be posed as a classification task with self-generated labels. The second general approach relies on auxiliary action labels, available in real or simulated robotic environments. These can either be used to train action-conditional predictive models of future frames \cite{Chiappa17,oh15} or inverse-kinematics models \cite{agrawal2016learning} which attempt to predict actions from current and future frame pairs. The third and most general approaches are predictive auto-encoders (e.g.\cite{hinton2006reducing,Kalchbrenner16,Mathieu15,Srivastava15}) which attempt to predict future frames from current ones. To learn effective representations, some kind of constraint on the latent representation is required.   

In this paper, we introduce a form of predictive auto-encoder which uses a novel adversarial loss to factor the latent representation for each video frame into two components, one that is roughly time-independent (i.e.~approximately constant throughout the clip) and another that captures the dynamic aspects of the sequence, thus varying over time. We refer to these as {\em content} and {\em pose} components, respectively. The adversarial loss relies on the intuition that while the content features should be distinctive of a given clip, individual pose features should not. Thus the loss encourages pose features to carry no information about clip identity. Empirically, we find that training with this loss to be crucial to inducing the desired factorization. 

We explore the disentangled representation produced by our model, which we call Disentangled-Representation Net (\drnet), on a variety of tasks. The first of these is predicting future video frames, something that is straightforward to do using our representation. We apply a standard LSTM model to the pose features, conditioning on the content features from the last observed frame. Despite the simplicity of our model relative to other video generation techniques, we are able to generate convincing long-range frame predictions, out to hundreds of time steps in some instances. This is significantly further than existing approaches that use real video data. 
We also show that \drnet can be used for classification. The content features capture the semantic content of the video thus can be used to predict object identity. Alternately, the pose features can be used for action prediction.

%% file: related.tex
\vspace{-6mm}
\section{Related work}
\vspace{-2mm}
On account of its natural invariances, image data naturally lends
itself to an explicit ``what'' and ``where'' representation. The
capsule model of Hinton \etal \cite{Hinton11} performed this
separation via an explicit auto-encoder structure. Zhao \etal
\cite{zhao2016} proposed a multi-layered version, which has
similarities to ladder networks \cite{ramus2015}. These methods all
operate on static images, whereas our approach uses temporal structure
to separate the components.

Factoring video into time-varying and time-independent components has
been explored in many settings. Classic structure-from-motion methods
use an explicit affine projection model to extract a 3D point cloud
and camera homography matrices \cite{hartley2000multiple}. In
contrast, Slow Feature Analysis \cite{Wiskott02} has no model, instead
simply penalizing the rate of change in time-independent components and
encouraging their decorrelation. Most closely related to ours is
Villegas \etal \cite{Villegas17} which uses an unsupervised approach to factoring
video into content and motion. Their architecture is also broadly similar to
ours, but the loss functions differ in important ways. They rely on
pixel/gradient space $\ell_p$-norm reconstructions, plus a GAN term
\cite{goodfellow2014} that encourages the generated frames to be sharp. 
We also use an $\ell_2$ pixel-space reconstruction. However, this pixel-space loss is only applied, in combination with a novel adversarial term applied to the pose features, to learn the disentangled representation.
In contrast to \cite{Villegas17}, our forward model acts on latent pose vectors rather than predicting pixels directly. 

Other approaches explore general methods for learning disentangled
representations from video. Kulkarni \etal \cite{kulkarni2015deep} show how explicit
graphics code can be learned from datasets with systematic dimensions
of variation. Whitney \etal \cite{whitney2016understanding} use a
gating principle to encourage each dimension of the latent
representation to capture a distinct mode of variation.


A range of generative video models, based on deep nets, have recently
been proposed. Ranzato \etal \cite{Ranzato14} adopt a discrete vector
quantization approach inspired by text models. Srivastava \etal \cite{Srivastava15} use
LSTMs to generate entire frames. Video Pixel Networks
\cite{Kalchbrenner16} use these models is a conditional manner,
generating one pixel at a time in raster-scan order (similar image
models include \cite{salimans2017pixelcnn++, Oord16}). Finn \etal \cite{Finn16}
use an LSTM framework to model motion via transformations of groups of
pixels.  Cricri \etal \cite{cricri2016video} use a ladder of stacked-autoencoders.
Other works predict optical flows fields that can be used to
extrapolate motion beyond the current frame,
e.g. \cite{liu09,visualdynamics16,walker15}. In contrast, a single
pose vector is predicted in our model, rather than a spatial field.
 
Chiappa \etal \cite{Chiappa17} and Oh \etal \cite{oh15} focus on prediction in video game
environments, where known actions at each frame can be permit  action-conditional generative models that can give accurate
long-range predictions. In contrast to the above works, whose latent
representations combine both content and motion, our approach relies
on a factorization of the two, with a predictive model only being
applied to the latter. Furthermore, we do not attempt to predict pixels
directly, instead applying the forward model in the latent
space.  Chiappa \etal \cite{Chiappa17}, like our approach, produces
convincing long-range generations. However, the video game environment
is somewhat more constrained than the real-world video we consider since actions are provided during generation.


Several video prediction approaches have been proposed that focus on
handling the inherent uncertainty in predicting the future. Mathieu
\etal \cite{mathieu2016} demonstrate the a loss based on GANs can produce sharper generations than traditional
$\ell_2$-based losses. \cite{Vondrick16} train a series of models,
which aim to span possible outcomes and select the most likely one at
any given instant. While we considered GAN-based losses, the more
constrained nature of our model, and the fact that our forward model does not directly
generate in pixel-space, meant that standard deterministic losses
worked satisfactorily.


%% file: approach.tex
\vspace{-3mm}
\section{Approach}
\vspace{-3mm}

In our model, two separate encoders produce distinct feature representations of content and pose for each frame. They are trained by requiring that the content representation of frame $x^t$ and the pose representation of future frame $x^{t+k}$ can be combined (via concatenation) and decoded to predict the pixels of future frame $x^{t+k}$. However, this reconstruction constraint alone is insufficient to induce the desired factorization between the two encoders. We thus introduce a novel adversarial loss on the pose features that prevents them from being discriminable from one video  to another, thus ensuring that they cannot contain content information. A further constraint, motivated by the notion that content information should vary slowly over time, encourages temporally close content vectors to be similar to one another.


More formally, let $x_i = (x_i^1, ..., x_i^T)$ denote a sequence of $T$ images from video $i$. We subsequently drop index $i$ for brevity. Let $E_c$ denote a neural network that maps an image $x^t$ to the \textit{content} representation $h_c^t$ which captures structure shared across time.  Let $E_p$ denote a neural network that maps an image $x^t$ to the \textit{pose} representation $h_p^t$ capturing content that varies over time. Let $D$ denote a decoder network that maps a content representation from a frame, $h_c^t$, and a pose representation $h_p^{t+k}$ from future time step $t+k$ to a prediction of the future frame $\tilde{x}^{t+k}$. Finally, $C$ is the {\em scene discriminator network} that takes pairs of pose vectors $(h_1,h_2)$ and outputs a scalar probability that they came from the same video or not.

The loss function used during training has several terms:

\textbf{Reconstruction loss:}
We use a standard per-pixel $\ell_2$ loss between the predicted future frame $\tilde{x}^{t+k}$ and the actual future frame $x^{t+k}$ for some random frame offset $k \in [0,K]$:
\begin{equation}
    \mathcal{L}_{reconstruction}(E_c, E_p, D) =  ||D(E_c(x^t), E_p(x^{t+k})) - x^{t+k}||^2_2
\end{equation}
Note that many recent works on video prediction that rely on more complex losses that can capture uncertainty, such as GANs \cite{mathieu2016, goodfellow2014}.

\textbf{Similarity loss:}
To ensure the content encoder extracts mostly time-invariant representations, we penalize the squared error between the content features $h_c^t, h_c^{t+k}$ of neighboring frames $k \in [0,K]$:
\begin{equation}
    \mathcal{L}_{similarity}(E_c) =  ||E_c(x^t) - E_c(x^{t+k}) ||^2_2
\end{equation}


\textbf{Adversarial loss:}
We now introduce a novel adversarial loss that exploits the fact that the objects present do not typically change {\em within} a video, but they do {\em between} different videos. Our desired disenanglement would thus have the content features be (roughly) constant within a clip, but distinct between them. This implies that the pose features should not carry any information about the identity of objects within a clip.

\begin{figure}
    \centering
\mbox{
    \includegraphics[width=0.4\linewidth]{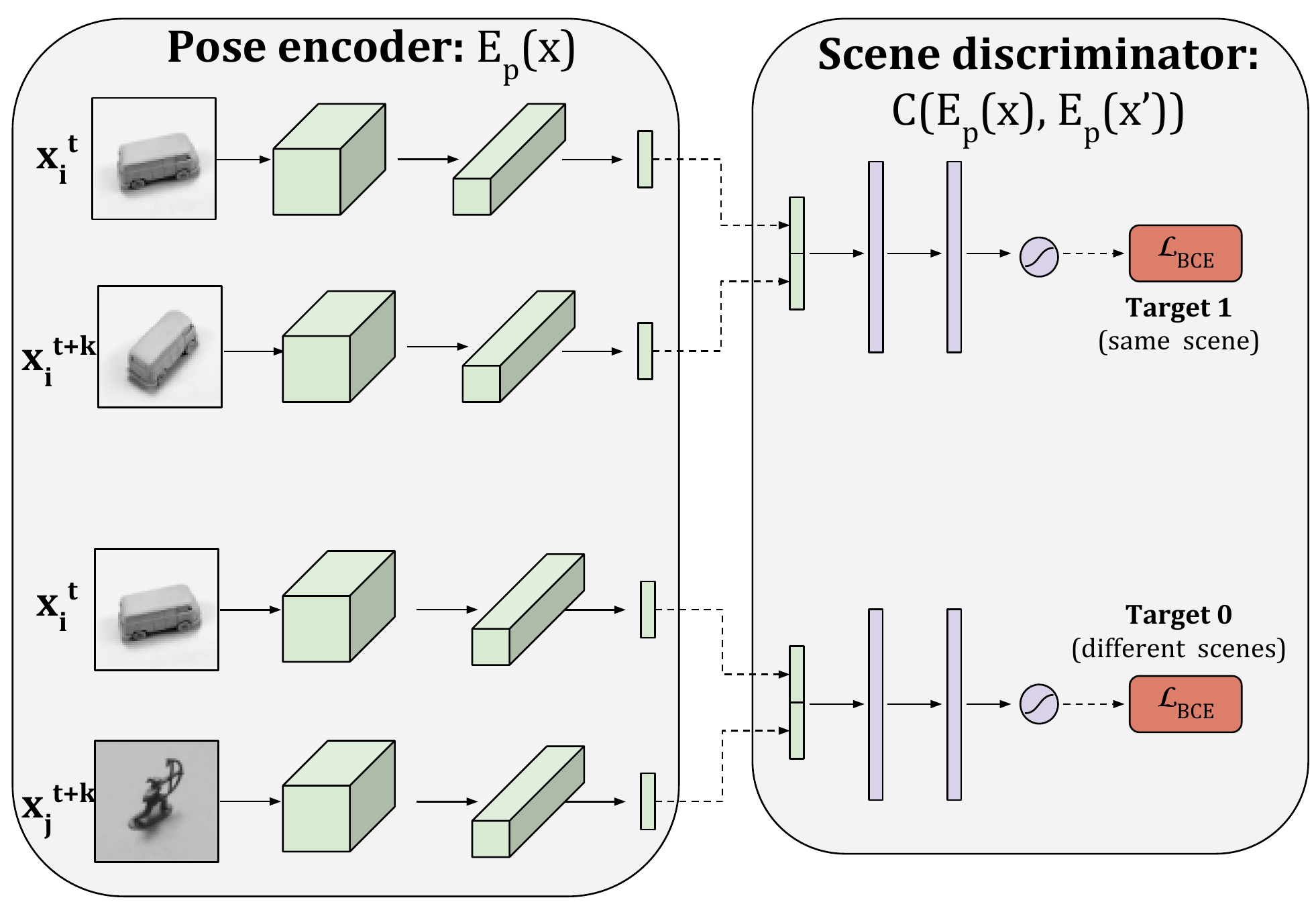}
\rulesep
  \includegraphics[width=0.55\linewidth]{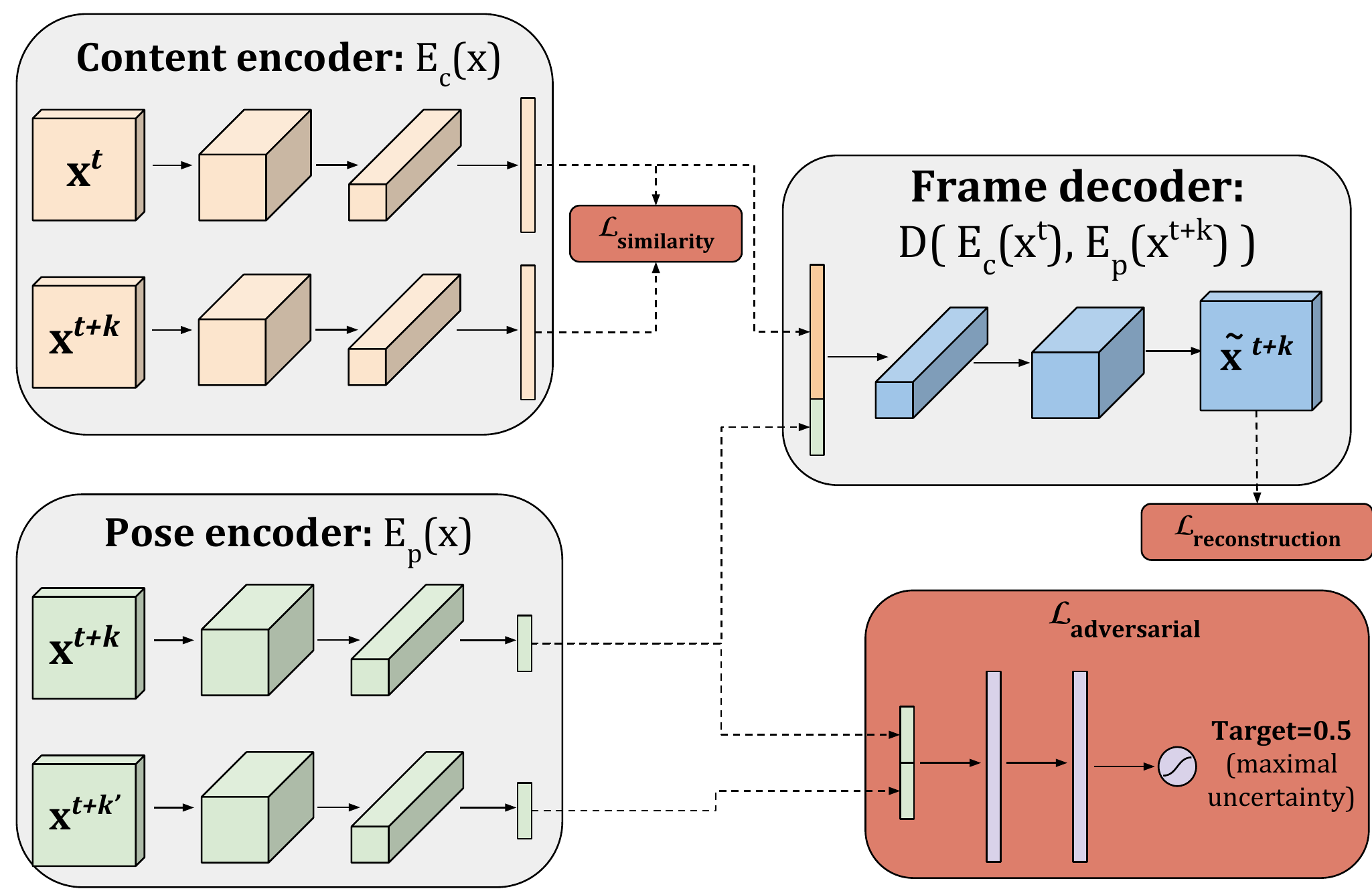}
 }
 \caption{Left: The discriminator $C$ is trained with binary cross entropy (BCE) loss to predict if a pair of pose vectors comes from the same (top portion) or different (lower portion) scenes. $x_i$ and $x_j$ denote frames from different sequences $i$ and $j$. The frame offset $k$ is sampled uniformly in the range $[0,K]$. Note that when $C$ is trained, the pose encoder $E_p$ is fixed. Right: The overall model, showing all terms in the loss function. Note that when the pose encoder $E_p$ is updated, the scene discriminator is held fixed.}
    \label{fig:model}
\end{figure}

We impose this via an adversarial framework between the scene discriminator network $C$ and pose encoder $E_p$, shown in \fig{model}. The latter provides pairs of pose vectors, either computed from the same video $(h_{p,i}^t, h_{p,i}^{t+k})$ or from different ones $(h_{p,i}^t, h_{p,j}^{t+k})$, for some other video $j$. The discriminator then attempts to classify the pair as being from the same/different video using a cross-entropy loss:
\begin{equation}
    - \mathcal{L}_{adversarial}(C) =  \log( C( E_p(x_i^t), E_p(x_i^{t+k})))  + \log( 1- C(E_p(x_i^t), E_p(x_j^{t+k})))
\end{equation}
The other half of the adversarial framework imposes a loss function on the pose encoder $E_p$ that tries to maximize the uncertainty (entropy) of the discriminator output on pairs of frames from the same clip:
\begin{equation}
    - \mathcal{L}_{adversarial}(E_p) = \frac{1}{2} \log( C( E_p(x_i^t), E_p(x_i^{t+k}))) + \frac{1}{2} \log( 1- C(E_p(x_i^t), E_p(x_i^{t+k})))
\end{equation}
Thus the pose encoder is encouraged to produce features that the discriminator is unable to classify if they come from the same clip or not. In so doing, the pose features cannot carry information about object content, yielding the desired factorization. Note that this does assume that the object's pose is not distinctive to a particular clip. While adversarial training is also used by GANs, our setup purely considers classification; there is no generator network, for example.


\textbf{Overall training objective:}\\
During training we minimize the sum of the above losses, with respect to $E_c, E_p, D$ and $C$:
\begin{equation}
\mathcal{L}  =
\mathcal{L}_{reconstruction}(E_c,E_p,D) + \alpha \mathcal{L}_{similarity}(E_c) + \beta (\mathcal{L}_{adversarial}(E_p) + \mathcal{L}_{adversarial}(C))
\end{equation}
where $\alpha$ and $\beta$ are hyper-parameters. The first three terms can be jointly optimized, but the discriminator $C$ is updated while the other parts of the model ($E_c,E_p,D$) are held constant. The overall model is shown in \fig{model}. Details of the training procedure and model architectures for $E_c, E_p, D$ and $C$ are given in \secc{model_details}.

\subsection{Forward Prediction}
\label{sec:forward}
After training, the pose and content encoders $E_p$ and $E_c$ provide a representation which enables video prediction in a straightforward manner. Given a frame $x^t$, the encoders produce $h^t_p$ and $h^t_c$ respectively. To generate the next frame, we use these as input to an LSTM model to predict the next pose features $h^{t+1}_p$. These are then passed (along with the content features) to the decoder, which generates a pixel-space prediction $\tilde{x}^{t+1}$:
\begin{eqnarray}
\tilde{h}^{t+1}_p &=& LSTM(E_p(x^t), h^t_c) \hspace{15mm}  \tilde{x}^{t+1} = D(\tilde{h}^{t+1}_p ,h^t_c) \\
\tilde{h}^{t+2}_p &=& LSTM(\tilde{h}^{t+1}_p, h^t_c) \hspace{18mm}  \tilde{x}^{t+2} = D(\tilde{h}^{t+2}_p ,h^t_c)
\end{eqnarray}
Note that while pose estimates are generated in a recurrent fashion, the content features $h^t_c$ remain fixed from the last observed real frame. This relies on the nature of $\mathcal{L}_{reconstruction}$ which ensured that content features can be combined with future pose vectors to give valid reconstructions.

The LSTM is trained separately from the main model using a standard $\ell_2$ loss between $\tilde{h}^{t+1}_p$ and $h^{t+1}_p$. Note that this generative model is far simpler than many other recent approaches, e.g.~\cite{Kalchbrenner16}. This largely due to the forward model being applied within our disentangled representation, rather than directly on raw pixels.

\begin{figure}
    \centering
    \includegraphics[width=0.9\linewidth]{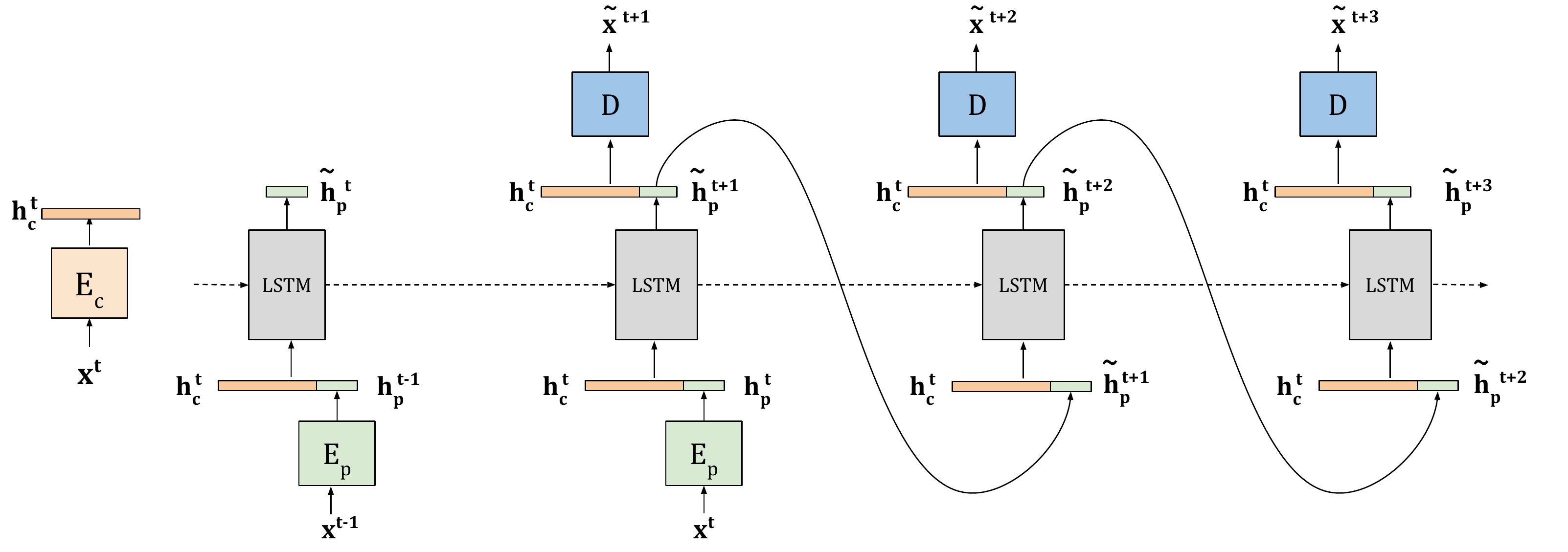}
    \caption{Generating future frames by recurrently predicting $h_p$, the latent pose vector. }
    \label{fig:prediction}
\vspace{-3mm}
\end{figure}

\subsection{Classification}
Another application of our disentangled representation is to use it for classification tasks. Content features, which are trained to be invariant to local temporal changes, can be used to classify the semantic content of an image. Conversely, a sequence of pose features can be used to classify actions in a video sequence. In either case, we train a two layer classifier network $S$ on top of either $h_c$ or $h_p$, with its output predicting the class label $y$.


%% file: experiments.tex
\section{Experiments}
We evaluate our model on both synthetic (MNIST, NORB, SUNCG) and real
(KTH Actions) video datasets. We explore several tasks with our model:
(i) the ability to cleanly factorize into content and pose components;
(ii) forward prediction of video frames using the approach from
\secc{forward}; (iii) using the pose/content features for
classification tasks.

\subsection{Model details} 
\label{sec:model_details}
We explored a variety of convolutional
architectures for the content encoder $E_c$, pose encoder $E_p$ and
decoder $D$.  For MNIST, $E_c,E_p$ and $D$ all use a DCGAN
 architecture \cite{radford2016} with $\left\vert{h_p}\right\vert = 5$
and $\left\vert{h_c}\right\vert = 128$.  The encoders consist of 5
convolutional layers with subsampling. Batch normalization and Leaky
ReLU's follow each convolutional layer except the final layer which 
normalizes the pose/content vectors to have unit norm.  The decoder is
a mirrored version of the encoder with 5 deconvolutional layers and a sigmoid output layer.

For both NORB and SUNCG, $D$ is a DCGAN architecture while $E_c$ and
$E_p$ use a ResNet-18 architecture \cite{resnet} up until the final pooling layer with  $\left\vert{h_p}\right\vert = 10$ and  $\left\vert{h_c}\right\vert = 128$.

For KTH, $E_p$ uses a ResNet-18 architecture with
$\left\vert{h_p}\right\vert = 5$.  $E_c$ uses the same
architecture as VGG16 \cite{vgg} up until the final pooling layer with
$\left\vert{h_c}\right\vert = 128$. The decoder is a mirrored
version of the content encoder with pooling layers replaced with
spatial up-sampling. In the style of U-Net \cite{ronneberger2015u}, we add skip connections from the content encoder
to the decoder, enabling the model to easily generate static
background features.  

In all experiments the scene discriminator $C$ is a fully connected
neural network with 2 hidden layers of 100 units.  We trained all our
models with the ADAM optimizer \cite{Kingma2015} and learning rate
$\eta=0.002$. We used $\beta = 0.1$ for MNIST, NORB and SUNCG and $\beta = 0.0001$ for KTH experiments.
We used $\alpha = 1$ for all datasets.

For future prediction experiments we train a two layer LSTM with 256
cells using the ADAM optimizer.  On MNIST, we train the model by
observing 5 frames and predicting 10 frames.  On KTH, we train the
model by observing 10 frames and predicting 10 frames.

\subsection{Synthetic datasets} 
\label{sec:toy}
\begin{figure}[b!]
\mbox{
\includegraphics[width=0.47\linewidth]{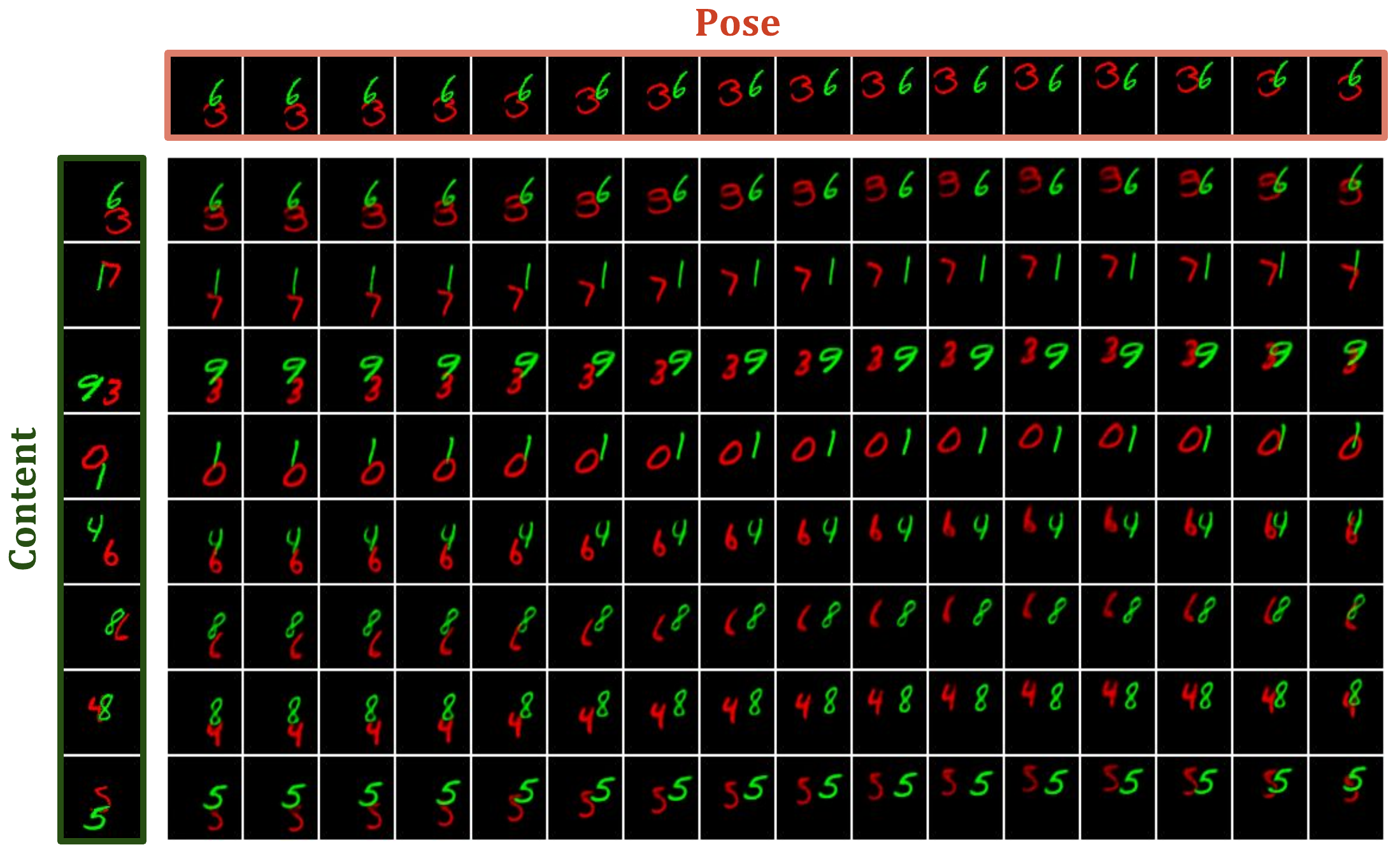}
\rulesep
\includegraphics[width=0.50\linewidth]{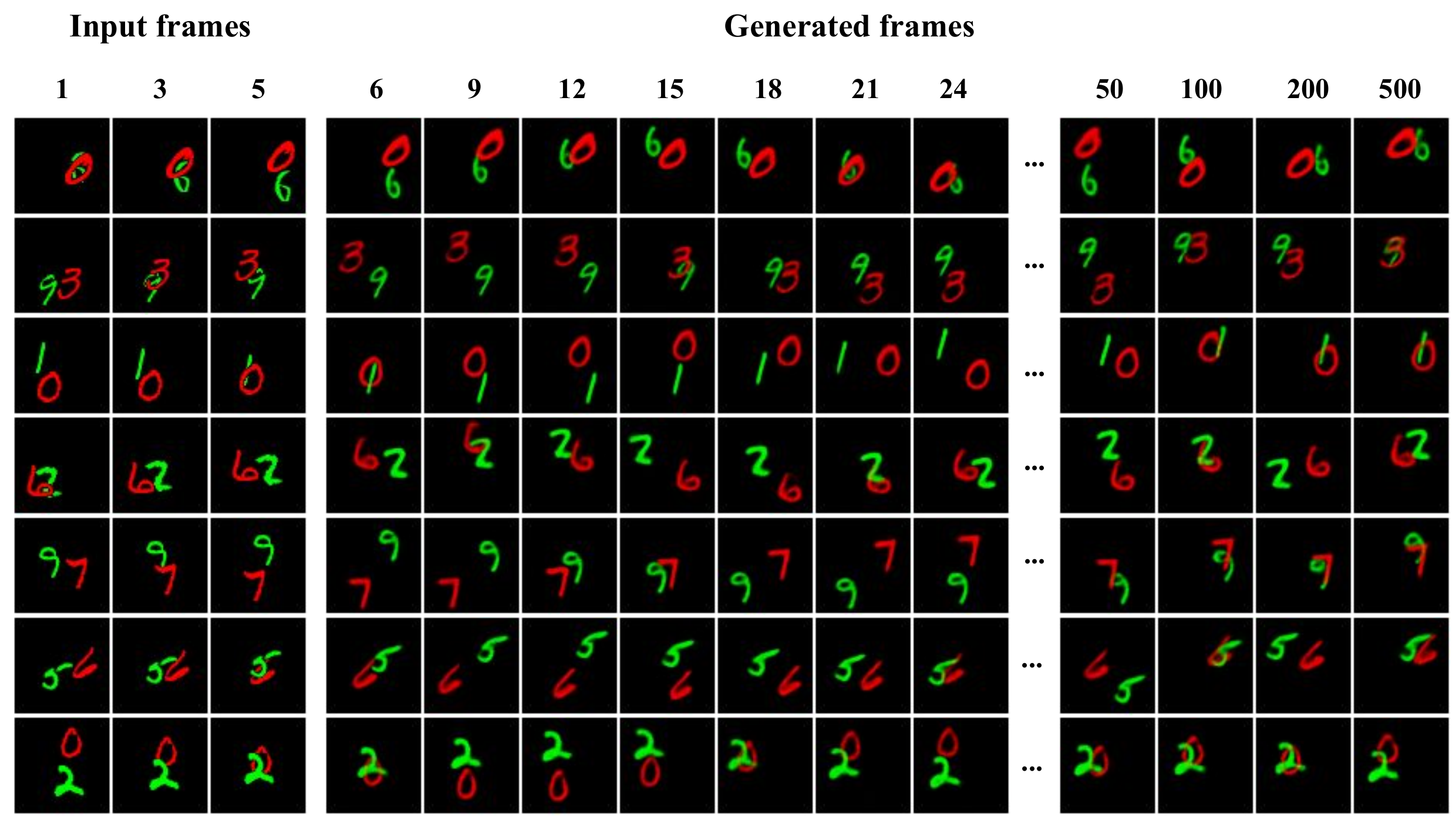}
}
\caption{Left: Demonstration of content/pose factorization on held out MNIST
  examples. Each image in the grid is
   generated using the pose and content vectors $h_p$ and $h_c$ taken
   from the corresponding images in the top row and first column
   respectively. The model has clearly
  learned to disentangle content and pose. Right: Each row shows
  forward modeling up to 500 time steps into the future, given 5
  initial frames. For each generation, note that only the pose
  part of the representation is being predicted from the previous time
  step (using an LSTM), with
  the content vector being fixed from the 5th frame. The
  generations remain crisp despite the long-range nature of
  the predictions. 
} \label{fig:mnist}
\end{figure}

{\bf MNIST:} We start with a toy dataset consisting of two MNIST
digits bouncing around a 64x64 image. Each video sequence consists of
a different pair of digits with independent
trajectories. \fig{mnist}(left) shows how the content vector from one
frame and the pose vector from another generate new examples that
transfer the content and pose from the original frames. This
demonstrates the clean disentanglement produced by our
model. Interestingly, for this data we found it to be necessary to
use a different color for the two digits. Our adversarial term is so
aggressive that it prevents the pose vector from capturing any content
information, thus without a color cue the model is unable to determine
which pose information to associate with which digit. 
In \fig{mnist}(right) we perform forward modeling using our
representation, demonstrating the ability to generate crisp digits 500
time steps into the future.    

{\bf NORB:} We apply our model to the NORB dataset \cite{norb}, converted into videos by taking sequences of different azimuths, while holding
object identity, lighting and elevation constant. \fig{norb}(left) shows
that our model is able to factor content and pose cleanly on held out
data. In \fig{norb}(center) we train a version of our model without
the adversarial loss term, which results in a significant
degradation in the model and the pose vectors are no longer isolated
from content. For comparison, we also show the factorizations produced
by Mathieu \etal \cite{mathieu2016}, which are less clean, both in
terms of disentanglement and generation quality than our
approach. Table 1 shows classification results on NORB, following the
training of a classifier on pose features and also content
features. When the adversarial term is used ($\beta=0.1$) the content
features perform well. Without the term, content features become less
effective for classification.    

\begin{figure}
    \includegraphics[width=\linewidth]{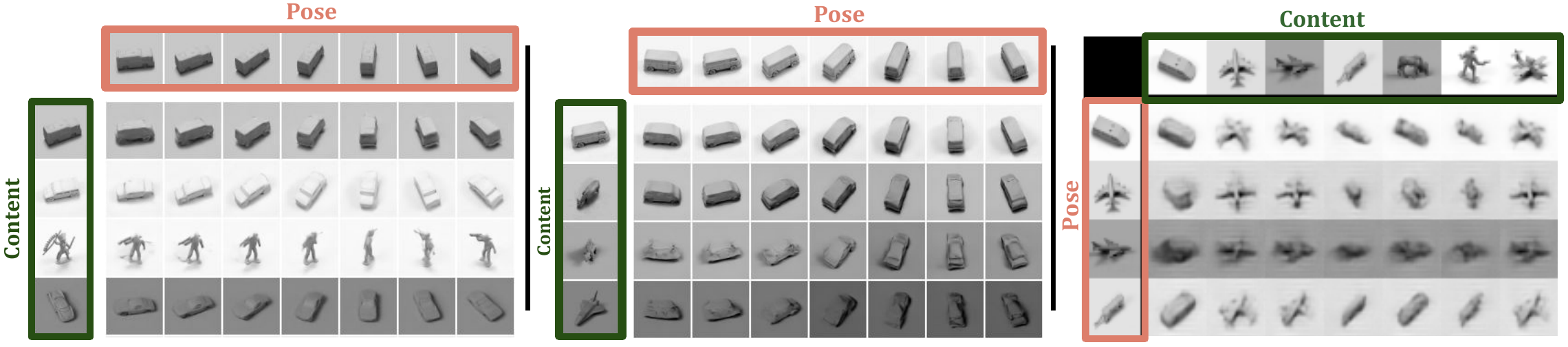}
\caption{Left: Factorization examples using
  our \drnet model on held out NORB images. Each image in the grid is
   generated using the pose and content vectors $h_p$ and $h_c$ taken
   from the corresponding images in the top row and first column
   respectively. Further examples can be found in
  the suplemental material. Center: Examples where \drnet was trained without
  the adversarial loss term. Note how content and pose are no longer
  factorized cleanly: the pose vector now contains
  content information which ends up dominating the generation. Right:
  factorization examples from Mathieu \etal \cite{mathieu2016}.} 
\end{figure}
\label{fig:norb}

{\bf SUNCG:} We use the rendering engine from the SUNCG dataset
\cite{song2016ssc} to generate sequences where the camera rotates
around a range of 3D chair models. \drnet is able to generate high
quality examples of this data, as shown in \fig{suncg}.
\begin{figure}
\mbox{
\begin{minipage}{0.3\textwidth}
   \includegraphics[width=1\linewidth]{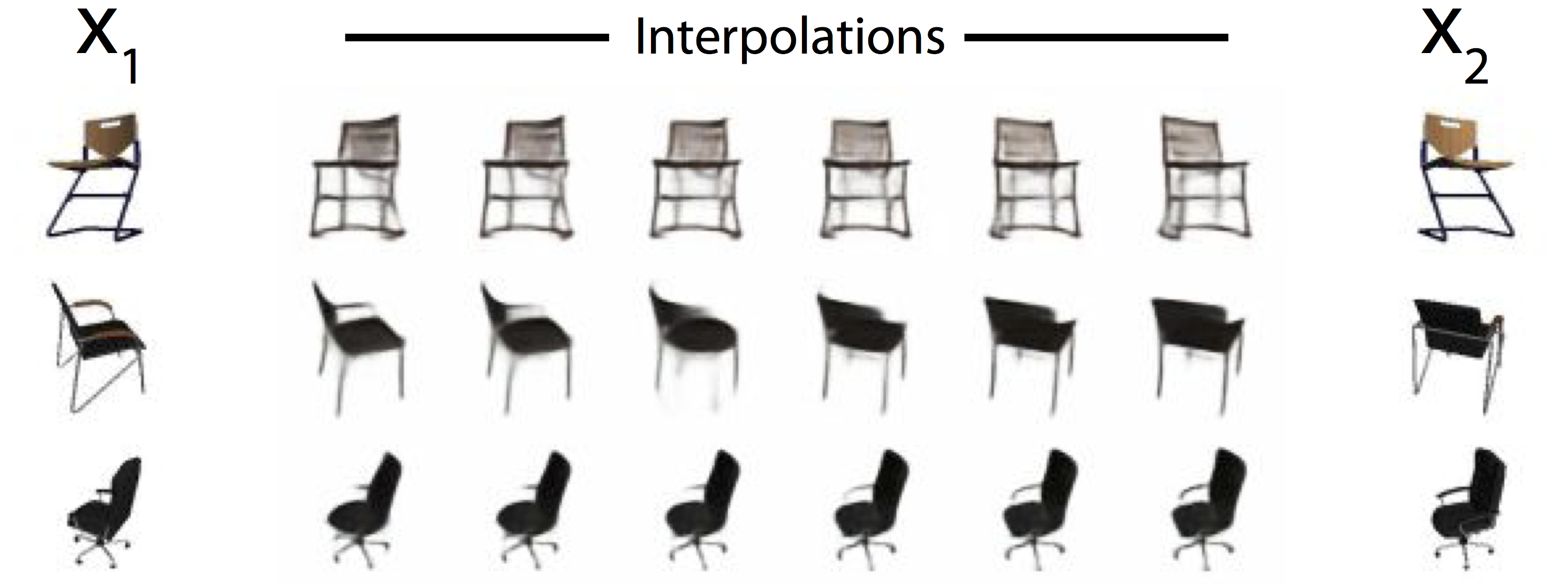}
   \includegraphics[width=1\linewidth]{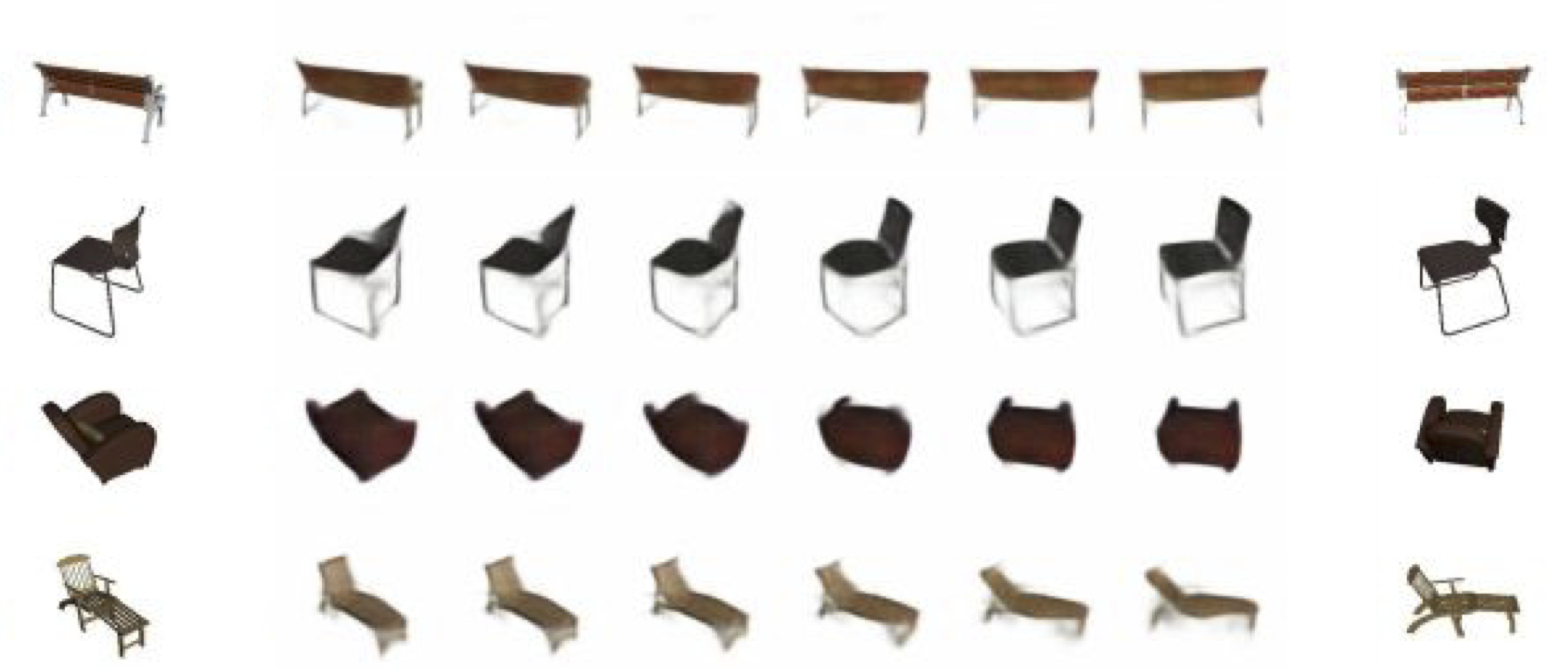}
   \includegraphics[width=1\linewidth]{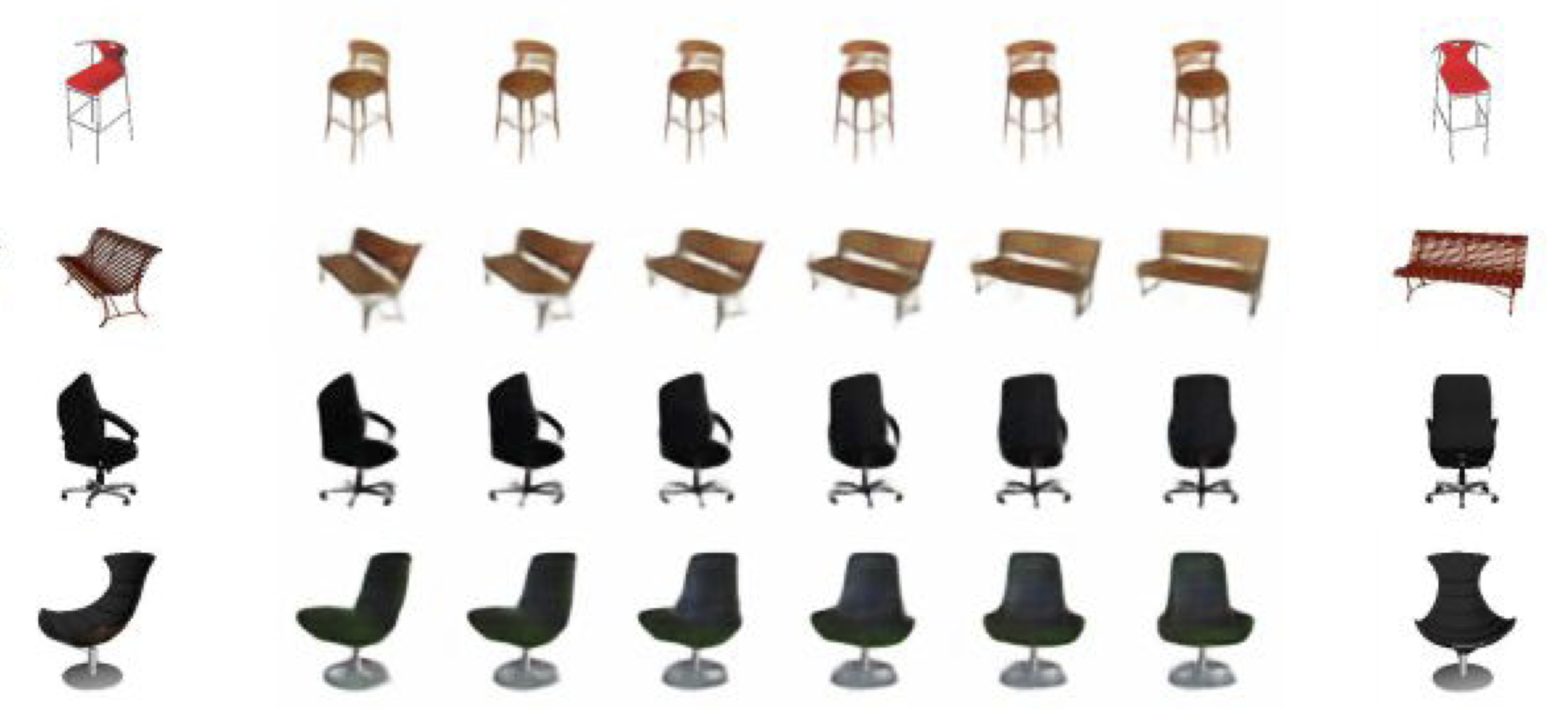}
\end{minipage}
\rulesep
\begin{minipage}{0.7\textwidth}
    \includegraphics[width=1\linewidth]{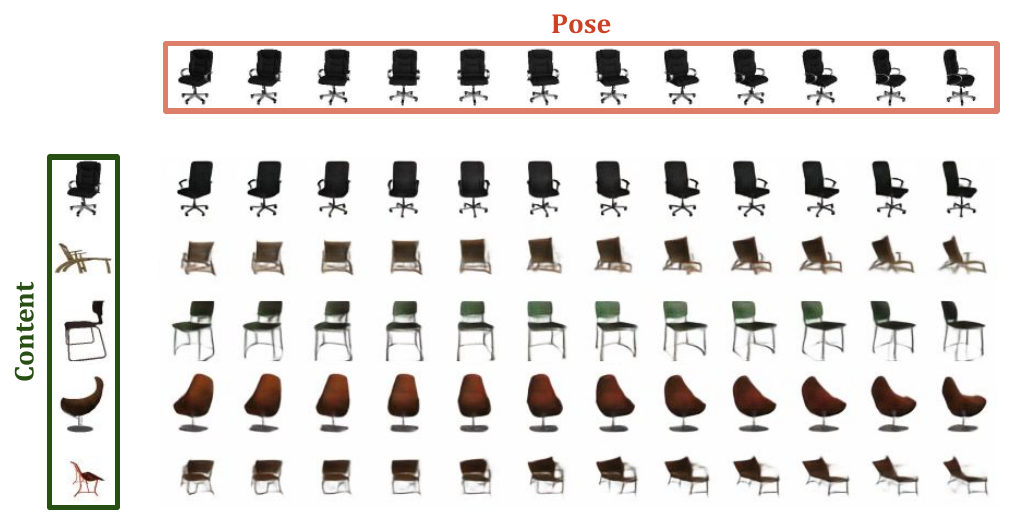}
\end{minipage}
 }
 \caption{Left: Examples of linear interpolation in pose space between
   the examples $x_1$ and $x_2$. Right: Factorization examples on held
   out images from the SUNCG dataset.  Each image in the grid is
   generated using the pose and content vectors $h_p$ and $h_c$ taken
   from the corresponding images in the top row and first column
   respectively. Note how, even for complex objects, the model is able
 to rotate them accurately. }
\end{figure}
\label{fig:suncg}

\vspace{-3mm}
\subsection{KTH Action Dataset}
\vspace{-2mm} 
Finally, we apply \drnet to the KTH dataset \cite{kth}. This is a 
simple dataset of real-world videos of people performing one of six
actions (walking, jogging, running, boxing, handwaving, hand-clapping)
against fairly uniform backgrounds. In \fig{kth-20} we show forward
generations of different held out examples, comparing against two
baselines: (i) the MCNet of Villegas \etal
\cite{Villegas17}
 which, to the best of our knowledge, produces the
current best quality generations of on real-world video and (ii) a
baseline auto-encoder LSTM model (AE-LSTM). This is essentially the
same as ours, but with a single encoder whose features thus combine
content and pose (as opposed to factoring them in
\drnet). It is also similar to \cite{Srivastava15}.

\begin{figure}[t!]
\begin{center}
   \includegraphics[width=\linewidth]{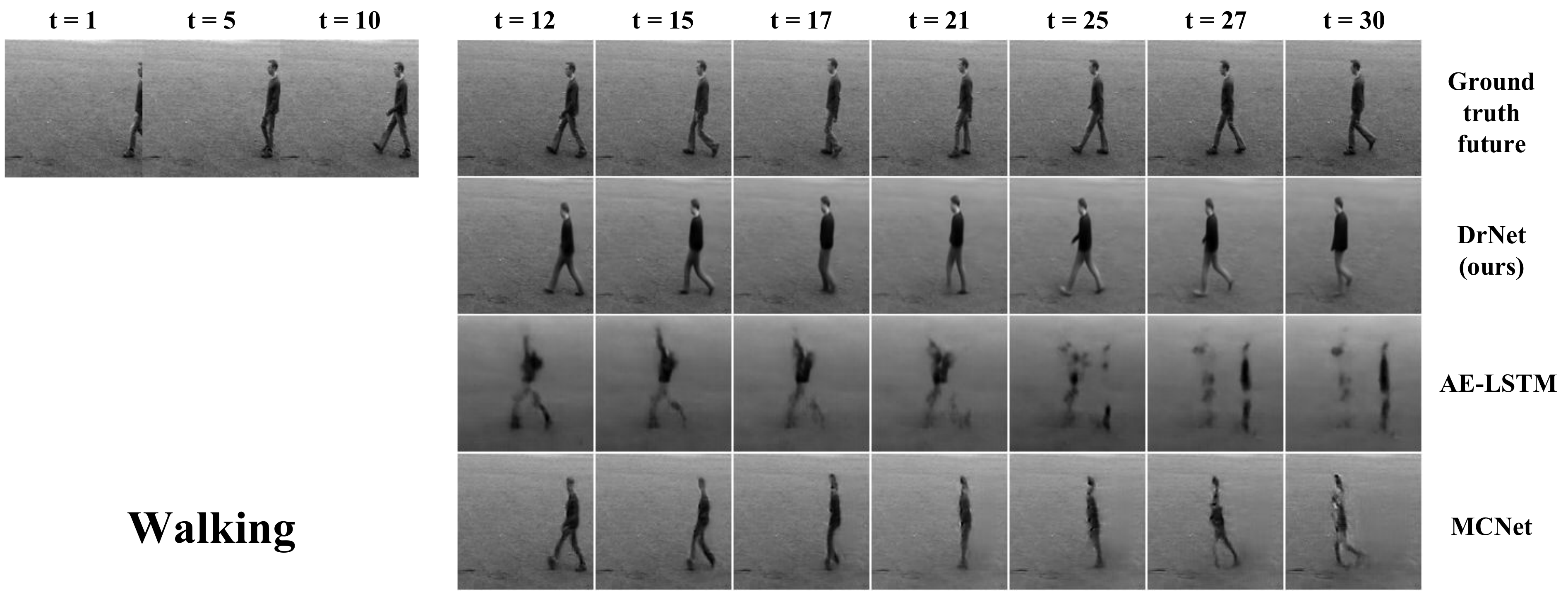}
    \includegraphics[width=\linewidth]{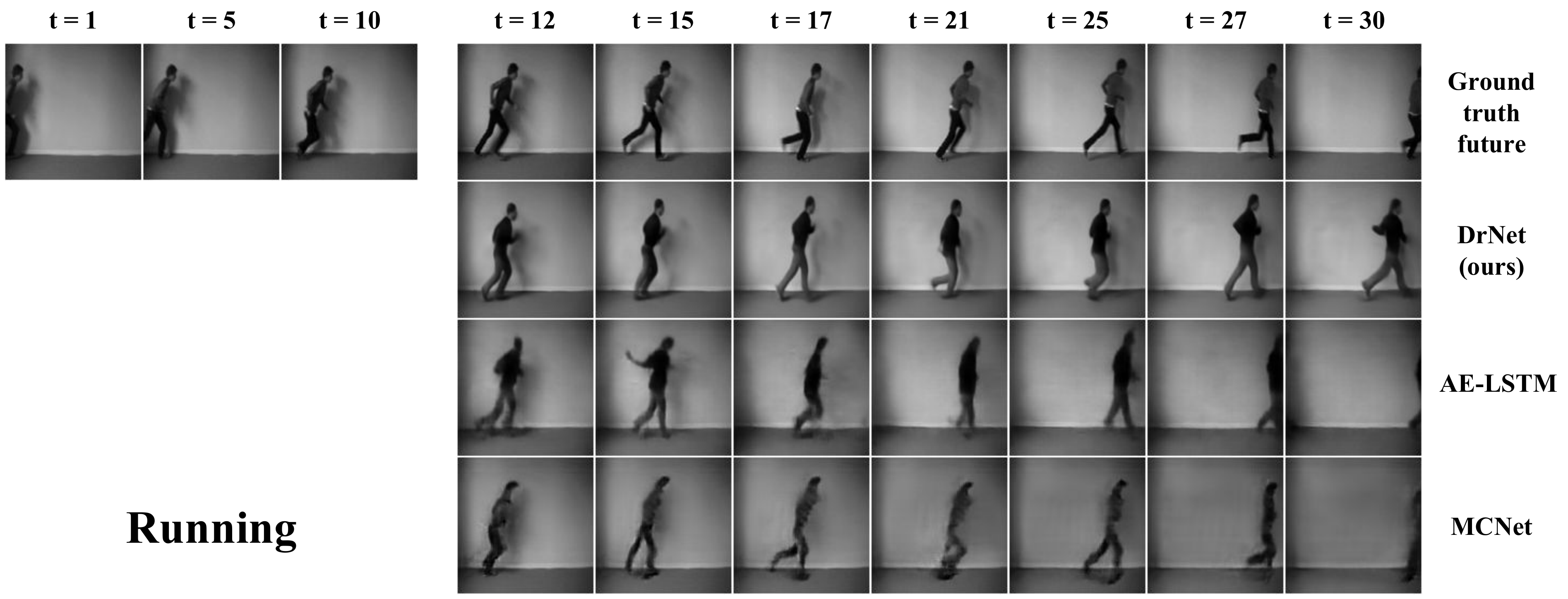}
\end{center}
\caption{Qualitative comparison between our \drnet model, MCNet
  \cite{Villegas17}  and the AE-LSTM baseline. All models are
  conditioned on the first 10 video frames and generate 20 frames. We
  display predictions of every 3$^{rd}$ frame. Video sequences are
  taken from held out examples of the KTH dataset for the classes of
  walking (top) and running (bottom).} 
\end{figure}
\label{fig:kth-20}

\begin{figure}[b!]
    \includegraphics[width=\linewidth]{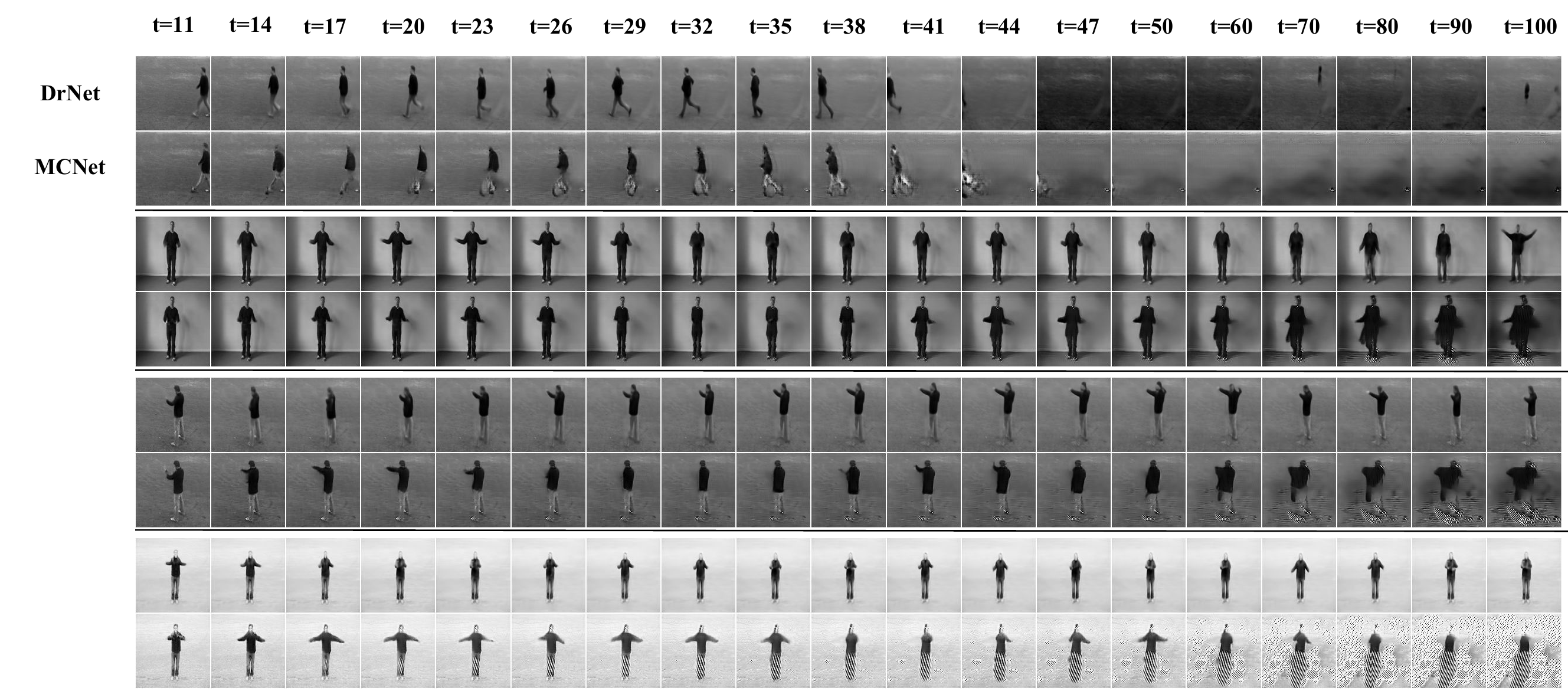}
\caption{Four additional examples of generations on held out examples
  of the KTH dataset, rolled out to 100 timesteps. } 
\label{fig:kth-100}
\end{figure}

\fig{kth-100} shows more examples, with generations out to
100 time steps. For most actions this is sufficient time for the person to
have left the frame, thus further generations would be of a fixed
background. In \fig{inception} we attempt to quantify the fidelity
of the generations by comparing the Inception score \cite{Salimans16}
of our approach to MCNet \cite{Villegas17}. This metric is used for
assessing generations from GANs and is more appropriate for our scenario
that traditional metrics such as PSNR or SSIM (see appendix \ref{inception} 
for further discussion). The curves show the mean scores of
our generations decaying more gracefully than MCNet \cite{Villegas17}. Further examples and
generated movies may be viewed in appendix \ref{moregen} and also at \url{https://sites.google.com/view/drnet-paper//}.

A natural concern with high capacity models is that they might be
memorizing the training examples. We probe this in \fig{kth_nn}, where
we show the nearest neighbors to our generated frames from the
training set. 
\fig{kth_class} uses the pose representation produced by
\drnet to train an action classifier from very few examples. 
We extract pose vectors from video sequences of length 24 and train a fully connected classifier on these vectors to predict the action class. 
We compare against an autoencoder baseline, which is the same as ours but with a single encoder whose features thus combine content and pose. 
We find the factorization significantly boosts performance.



\begin{minipage}{0.32\textwidth} 
    \centering
\tiny
    \begin{tabular}{ c c | c }
Model & &Accuracy (\%) \\
\hline
\hline
\multirow{2}{*}{\drnet$_{\beta=0.1}$} & $h_c$ & {\bf 93.3} \\
& $h_p$ & 60.9 \\ 
\hline
\multirow{2}{*}{\drnet$_{\beta=0}$} & $h_c$ & 72.6 \\
& $h_p$ & 80.8 \\ 
\hline
Mathieu \etal \cite{mathieu2016} &  & 86.5\\ 
\end{tabular}
\vspace{0mm}
    \captionof{table}{\small Classification results on NORB dataset,
      with/without adversarial loss ($\beta=0.1/0$) using content or pose
      representations ($h_c,h_p$ respectively). The adversarial term
      is crucial for forcing semantic information into the content
      vectors -- without it performance drops significantly. }
    \label{tab:norb_classify}
\end{minipage}
\hspace{2mm}
\rulesep
\hspace{2mm}
\begin{minipage}{0.27\textwidth} 

    \includegraphics[width=\linewidth]{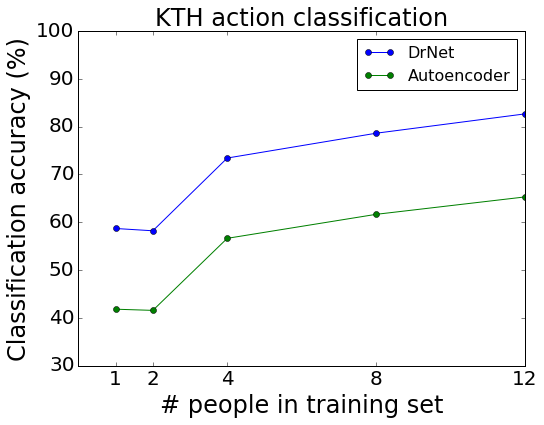}
\captionof{figure}{\small Classification of KTH actions from pose vectors with
  few labeled examples, with autoencoder baseline. N.B. SOA (fully
  supervised) is 93.9\% \cite{Le11}.}
\label{fig:kth_class}
\end{minipage}
\hspace{2mm}
\rulesep
\hspace{2mm}
\begin{minipage}{0.27\textwidth} 
    \includegraphics[width=\linewidth]{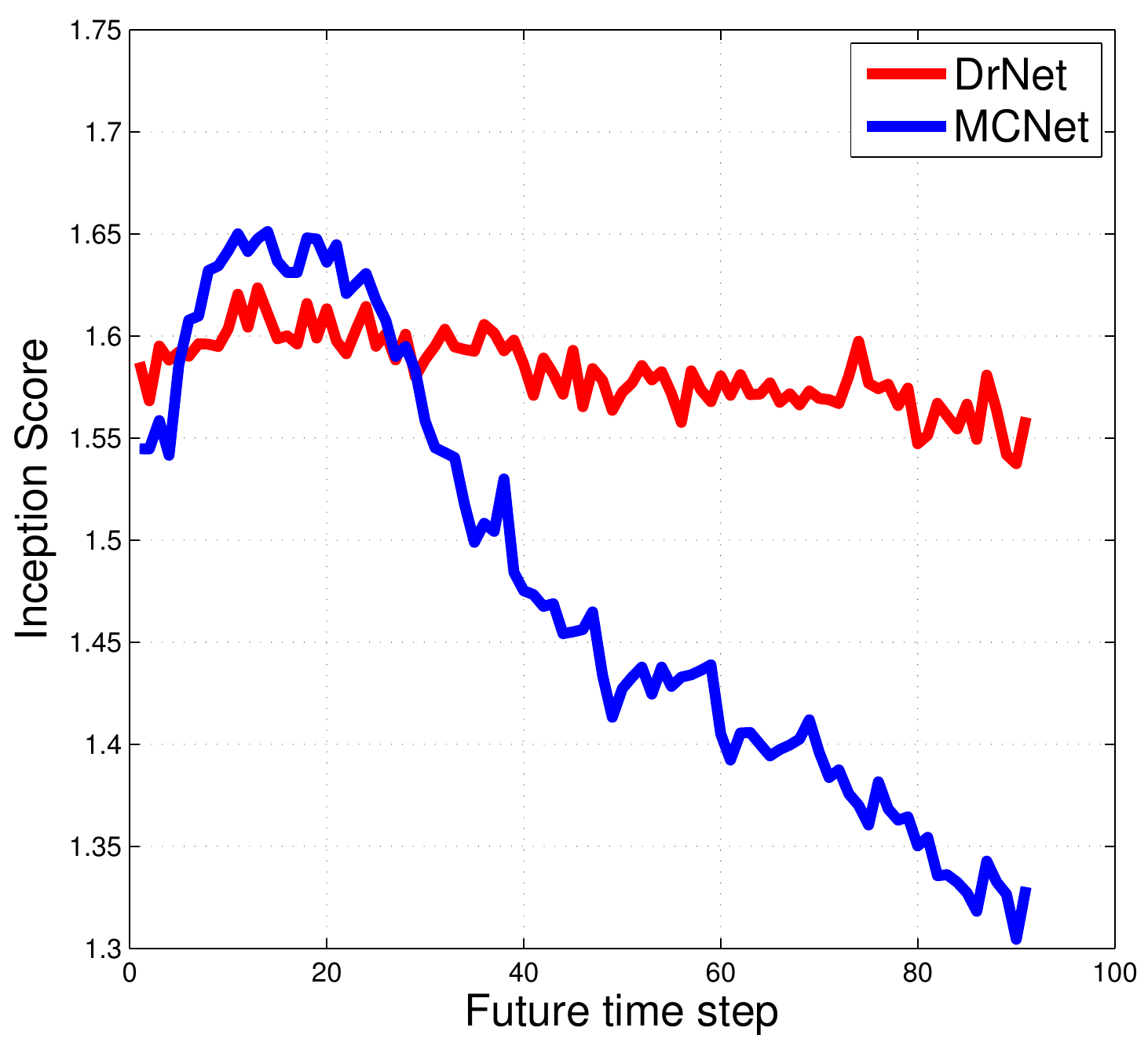}
    \captionof{figure}{\small Comparison of KTH video generation quality
      using Inception score.
X-axis indicated how far from conditioned input the start of the generated sequence is. 
}
\label{fig:inception}
\end{minipage}

\begin{figure}[h!]
    \includegraphics[width=\linewidth]{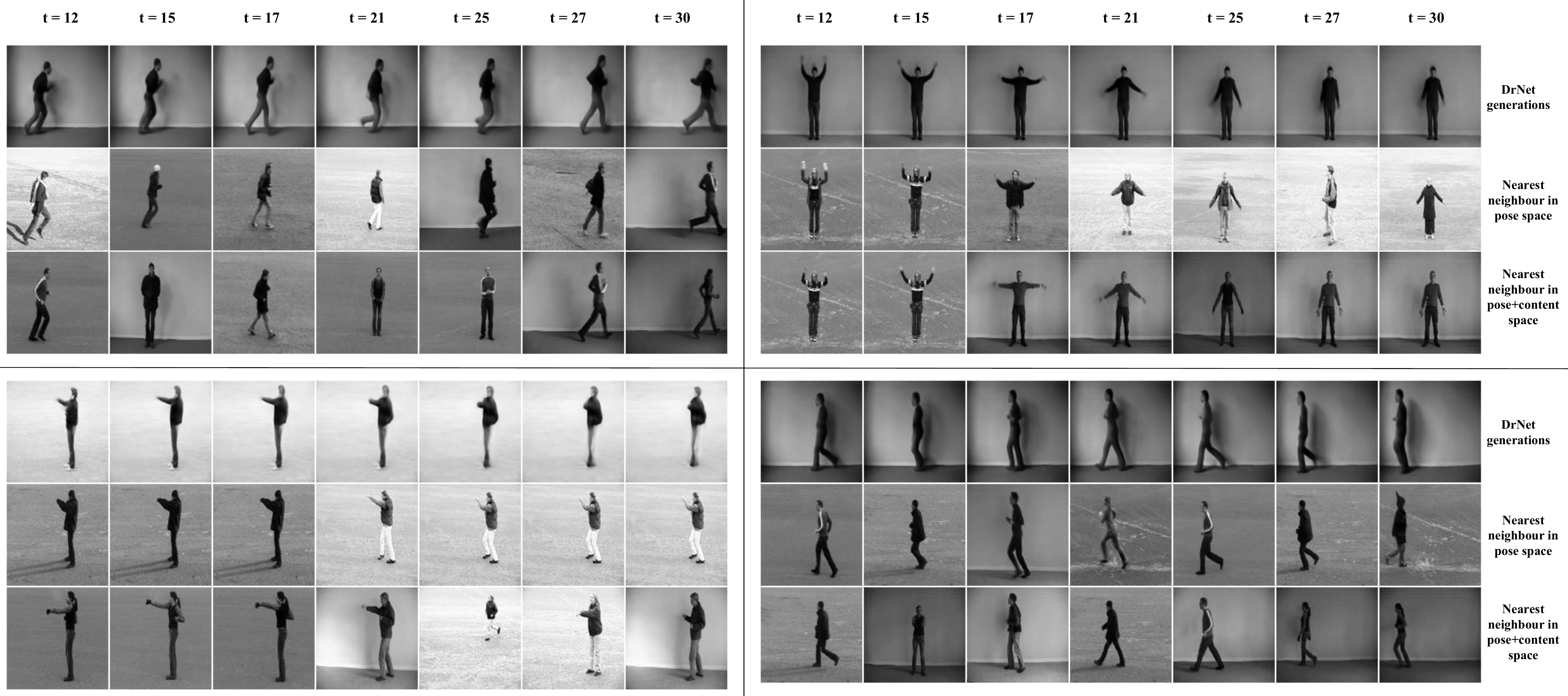}
\caption{For each frame generated by \drnet (top row in each set), we show
  nearest-neighbor images from the training set, based on pose vectors
(middle row) and both content and pose vectors (bottom row). It is
evident that our model is not simply copying examples from the
training data. Furthermore, the middle row shows that the pose vector
generalizes well, and is independent of background and clothing.} 
\end{figure}
\label{fig:kth_nn}

%% file: supp.tex
\clearpage
\appendix


\begin{figure}[b!]
\begin{center}
   \includegraphics[width=\linewidth]{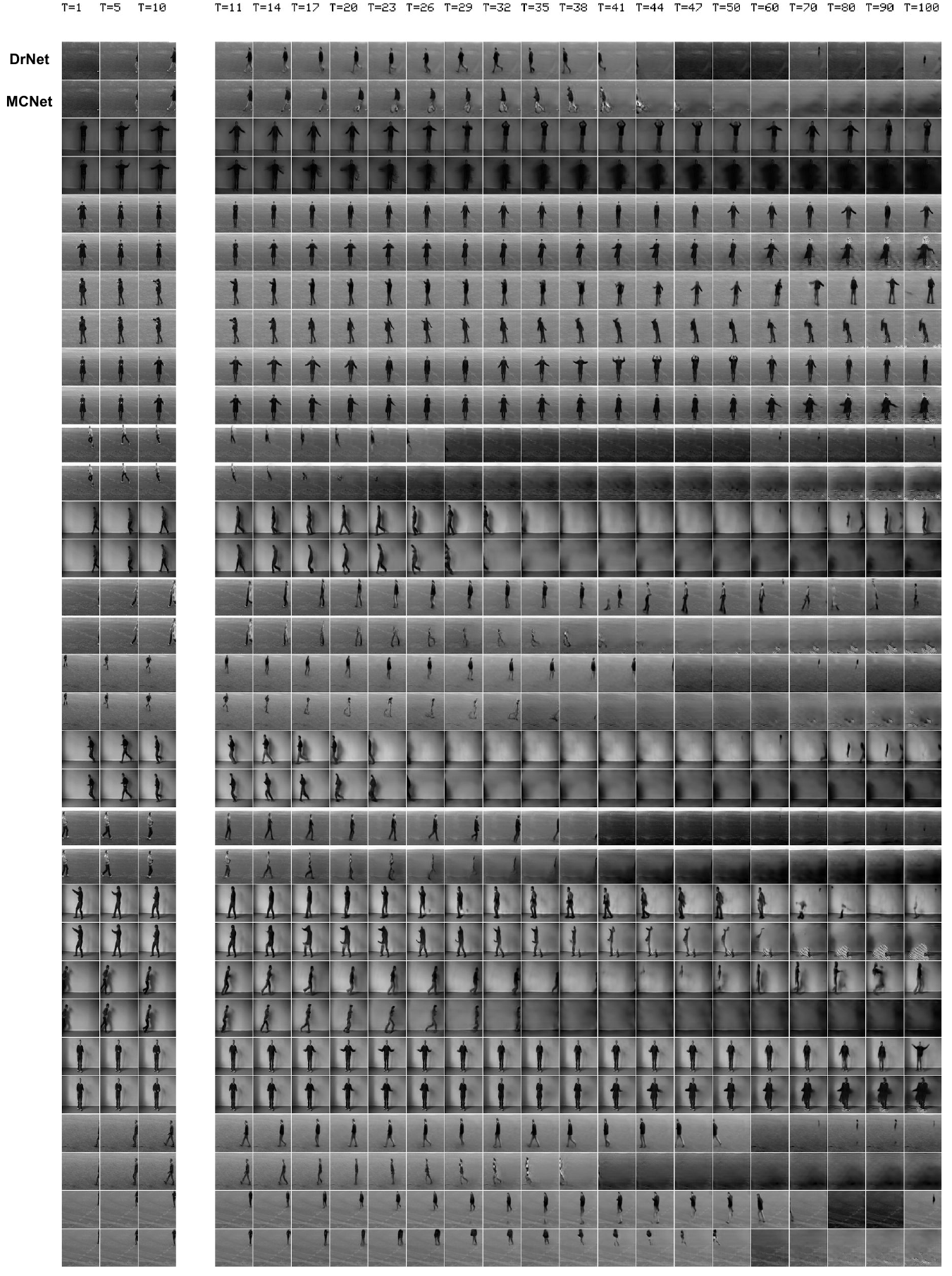}
\end{center}
\caption{Further long range generations on KTH dataset. Each pair of rows shows generations by \drnet (top) and MCNet (bottom) for the same conditioning frames.} 
\label{fig:kth100gens}
\end{figure}

\begin{figure}[b!]
\center
    \includegraphics[width=0.9\linewidth]{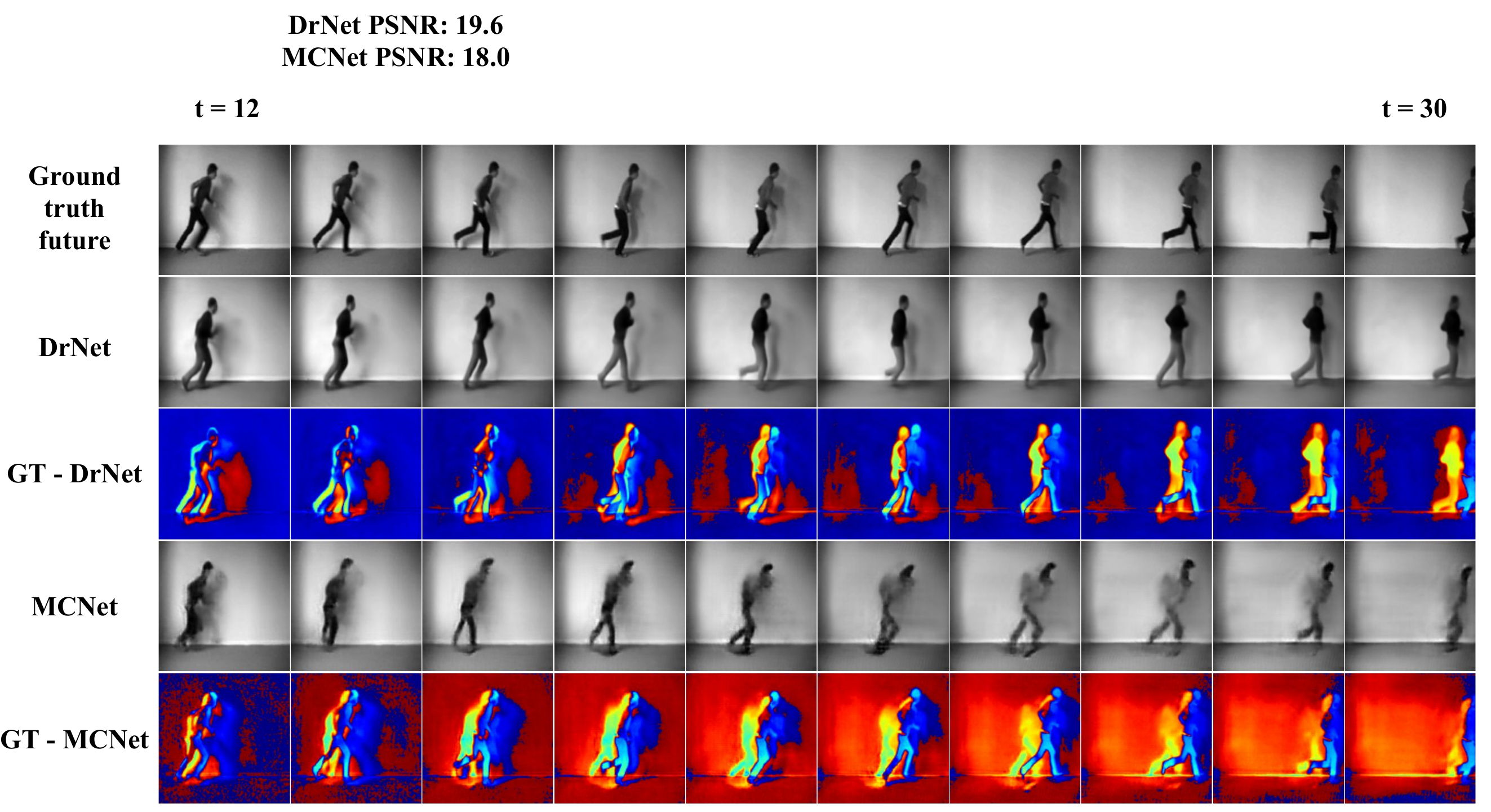}
\vspace{2mm}
\hrule
\vspace{2mm}
    \includegraphics[width=0.9\linewidth]{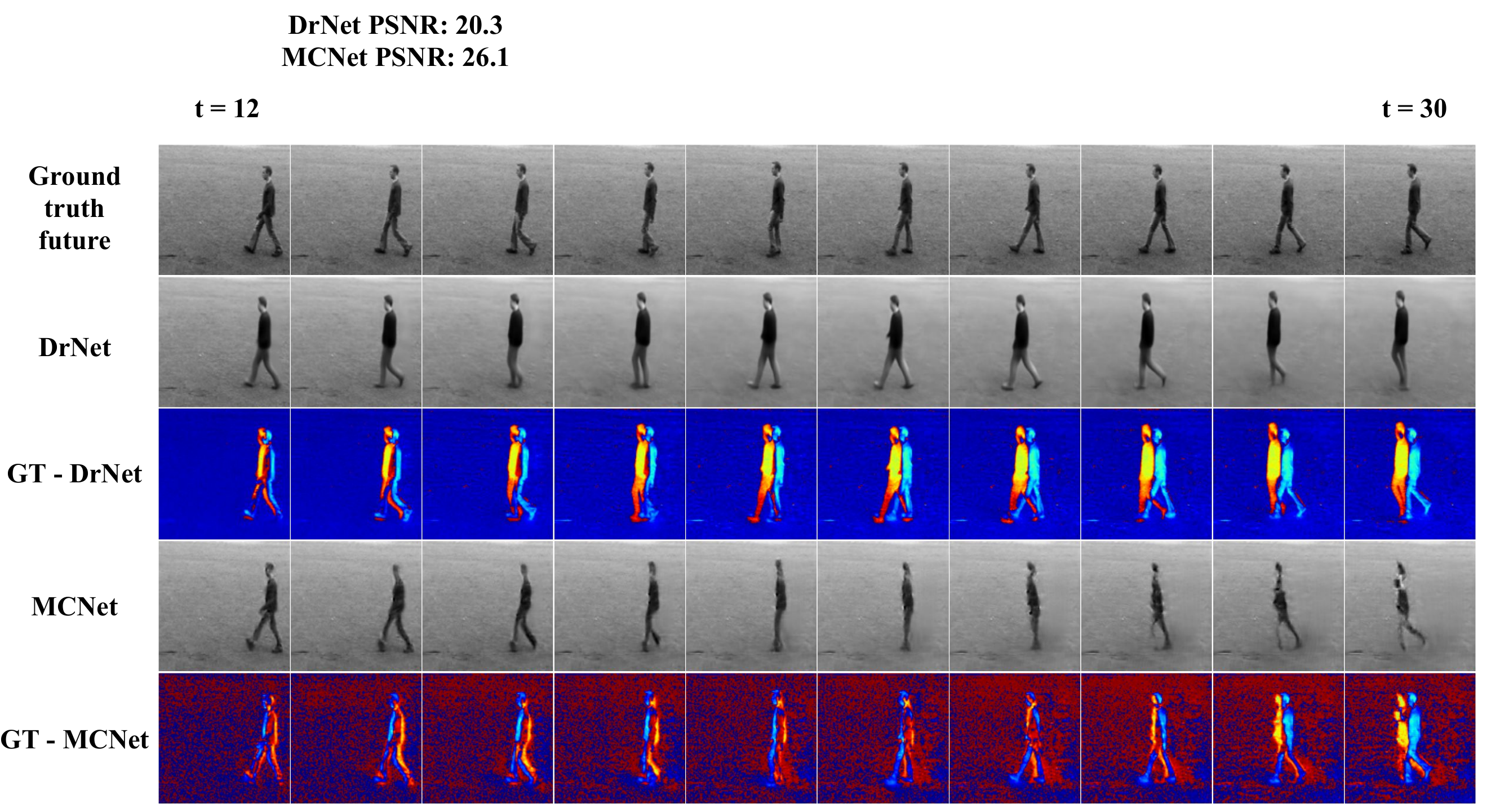}
\caption{Comparison of generated sequences from \drnet and MCNet along with corresponding PSNR scores.} 
\label{fig:psnr_ex1} 
\end{figure}

\section{Further KTH generations}
\label{moregen}
\fig{kth100gens} shows additional long-range KTH sequences generated from our model and MCNet \cite{Villegas17}.
Generations in movie form are viewable at
\url{https://sites.google.com/view/drnet-paper/}. 

\section{Quantitative metrics for evaluating generations}
\label{inception}
Evaluating
samples from generative models is generally problematic. Pixel-wise
measures like PNSR and SSIM \cite{wang2004image} are appropriate when
objects are well aligned, but for long-range generations this is
unlikely to be the case. \fig{psnr_ex1} shows sequences generated from
our model and MCNet \cite{Villegas17}, as well as their difference
with respect to the ground truth. While the person remains sharp, there is
some error in the velocity prediction, which accumulates to a
significant offset in position. Consequently, the resulting PSNR and
SSIM scores are very misleading and we adopt the Inception score
\cite{Salimans16} as an alternative in the main paper.

The Inception Score is computed by first training a classifier network to
accurately predict the action class of a video from a sequence of 10 frames. We employ a convolutional network classification architecture 
where the video frames are concatenated as input to the first layer. 
Once the classifier is trained, we evaluate the samples generated by \drnet and MCNet by considering the label distribution predicted by the classifier for generated sequences.
Intuitively, we expect a good generative model to produce videos with a highly peaked conditional label distribution $p(y|\mathbf{x})$ and a uniform marginal label distribution $p(y)$. 
Formally, the Inception Score is given by $\exp(\mathbb{E}_x \text{KL}(p(y|\mathbf{x}) ||p(y))$.
\fig{inception} plots the mean Inception Score for generated sequences from our \drnet model and MCnet \cite{Villegas17}.
The x-axis indicates the offset, from the final frame ofthe  conditioned input, of the first generated frame used for the Inception Score.

\section{KTH experimental settings}
The KTH dataset consists of 25 different subjects performing six different actions (boxing, hand waving, hand clapping, jogging, running and walking) against a static background. Each person is observed performaing every action in four different scenarios with varied clothing a background conditions. 
Following \cite{Villegas17} we used person 1-16 for training and person 17-25 for testing\footnote{Note, this is not the standard train/test split used for KTH action classification}. We also resize frames to 128$\times$128 pixels. 

\section{Details of classification experiments}
\label{sec:kth_details}
\textbf{NORB object classification:} We used a two layer fully connected network with 256 hidden units as the classifier.  
Leaky ReLU's, batch normalization and dropout were used in every layer.
We trained with ADAM as used early stopping on a validation set to prevent over fitting.

\textbf{KTH action classification:} 
We used a two layer fully connected network with 1200 hidden units as the classifier.  
Leaky ReLU's, batch normalization and dropout were used in every layer.
Both \drnet and the autoencoder baseline produced 24 dimensional latent vectors.
The classifier was trained on sequences of length 24 so the input to the classifier was 24$\times$24.
We also tried an autoencoder baseline with a 128 dimensional latent space (i.e, same dimensionality as the content vectors of \drnet) but found this model performed worse.
We trained the action classifier on the pose representations from \drnet and the autoencoder for varying training set sizes. 
Specifically, we varied the number of subjects used in the training set from 1 to 12.


\small